\documentclass[letterpaper]{article} 
\usepackage{aaai23}  
\usepackage{times}  
\usepackage{helvet}  
\usepackage{courier}  
\usepackage[hyphens]{url}  
\usepackage{graphicx} 
\usepackage{natbib}  
\usepackage{caption} 
\usepackage{algorithm}
\usepackage{algpseudocode} 
\usepackage{amsfonts}
\usepackage{amsmath}
\usepackage{multirow}
\usepackage{array}
\usepackage{booktabs}

\usepackage{newfloat}
\usepackage{listings}
\DeclareCaptionStyle{ruled}{labelfont=normalfont,labelsep=colon,strut=off} 
\floatstyle{ruled}
\newfloat{listing}{tb}{lst}{}
\floatname{listing}{Listing}
\title{Hierarchical Text Classification as Sub-hierarchy Sequence Generation}
\author{
    SangHun Im,
    GiBaeg Kim,
    Heung-Seon Oh,
    Seongung Jo,
    Dong Hwan Kim
}
\affiliations{
    School of Computer Science and Engineering, 
    Korea University of Technology and Education (KOREATECH)\\
    \{tkrhkshdqn, fk0214, ohhs, oowhat, hwan6615\}@koreatech.ac.kr
}

\begin{document}

\maketitle

\begin{abstract}
Hierarchical text classification (HTC) is essential for various real applications. However, HTC models are challenging to develop because they often require processing a large volume of documents and labels with hierarchical taxonomy. Recent HTC models based on deep learning have attempted to incorporate hierarchy information into a model structure. Consequently, these models are challenging to implement when the model parameters increase for a large-scale hierarchy because the model structure depends on the hierarchy size. To solve this problem, we formulate HTC as a sub-hierarchy sequence generation to incorporate hierarchy information into a target label sequence instead of the model structure. Subsequently, we propose the \textbf{Hi}erarchy \textbf{DEC}oder (HiDEC), which decodes a text sequence into a sub-hierarchy sequence using recursive hierarchy decoding, classifying all parents at the same level into children at once. In addition, HiDEC is trained to use hierarchical path information from a root to each leaf in a sub-hierarchy composed of the labels of a target document via an attention mechanism and hierarchy-aware masking. HiDEC achieved state-of-the-art performance with significantly fewer model parameters than existing models on benchmark datasets, such as RCV1-v2, NYT, and EURLEX57K.
\end{abstract}

\section{Introduction}

Hierarchical text classification (HTC) uses a hierarchy such as a web taxonomy to classify a given text into multiple labels. Moreover, classification tasks are essential in real-world applications because of the tremendous amount of data on the web that should be properly organized for applications such as product navigation  \cite{product1,product2} and news categorization \cite{rcv,nyt}.

Recent HTC research using deep learning can be categorized into local and global approaches. In the local approach \cite{local_dgcnn,local_hdltex,local_hft,local_htrans,local_parent,local_hmcn}, a classifier is built for each unit after the entire hierarchy is split into a set of small units. Subsequently, the classifiers are applied in sequence according to a path from a root to target labels in a top-down manner. In contrast, in the global approach \cite{global_capsule,global_hcsm,global_heagrcnn,global_hgclr,global_hiagm,global_hilap,global_himatch,global_hnatc,global_htcinfomax,global_l2l,global_sgm}, a classifier for all labels in the entire hierarchy is built, excluding the hierarchy structure through flattening. A document hierarchy information can be obtained using a structure encoder and merged with text features from a text encoder. The global approach achieves superior performance to the local approach owing to the effective design of the structure encoders, such as the graph convolution network \cite{gcn} and Graphormer \cite{graphormer}.

\begin{figure}[t]
\centering
\includegraphics[width=0.9\columnwidth]{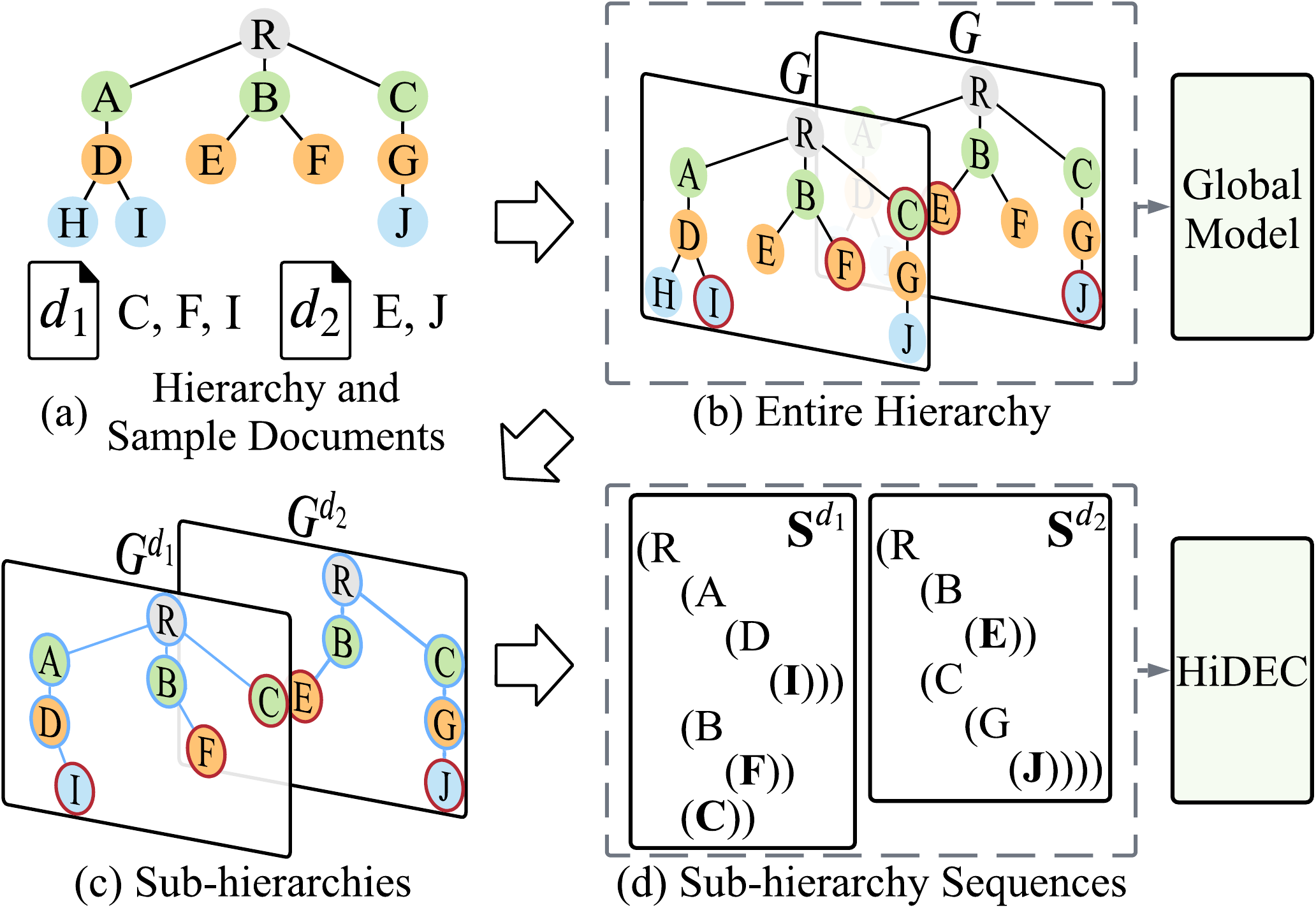} 
\caption{Example of converting the target labels of two documents to the sub-hierarchy sequences. The existing global model uses the entire hierarchy twice. In contrast, the proposed HiDEC uses the sufficiently small sub-hierarchies relevant to the documents twice.}
\label{fig1}
\end{figure}

However, the global approach has scalability limitations because the structure encoders require a disproportionate number of parameters as the size of the hierarchies increases. For example, HiAGM \cite{global_hiagm} and HiMatch \cite{global_himatch}, recent GCN-based structure encoders, require a weight matrix to convert text features to label features. In HGCLR \cite{global_hgclr}, edge and spatial encodings were employed to represent the relationship between two nodes.

In existing global models, a large model size is inevitable because they attempt to incorporate the entire hierarchy information with respect to the document labels into a model structure. In contrast, employing a sub-hierarchy comprising a set of paths from a root to each target label is sufficient because most labels are irrelevant to a target document. To this end, we formulate HTC as a sub-hierarchy sequence generation using an encoder-decoder architecture to incorporate the sub-hierarchy information into a target label sequence instead of the model structure. Figure \ref{fig1} shows the differences between the proposed approach and the existing global models. For example, given a hierarchy and two documents with different labels in Figure \ref{fig1}-(a), a global model attempts to capture the entire hierarchy information with respect to the document labels, as shown in Figure \ref{fig1}-(b). For further improvement, the document labels are converted into a sub-hierarchy sequence using a depth-first search on an entire hierarchy and parse tree notation, as shown in Figure \ref{fig1}-(c) and -(d).

Based on this idea, we propose a \textbf{Hi}erarchy \textbf{DEC}oder (HiDEC)\footnote{Code is available on https://github.com/SangHunIm/HiDEC}, which recursively decodes the text sequence into a sub-hierarchy sequence by sub-hierarchy decoding while remaining aware of the path information. The proposed method comprises hierarchy embeddings, hierarchy-aware masked self-attention, text-hierarchy attention, and sub-hierarchy decoding, similar to the decoder of Transformer \cite{transformer}. Hierarchy-aware masked self-attention facilitates learning all hierarchy information in a sub-hierarchy sequence. A hierarchy-aware mask captures the sub-hierarchy information by considering the dependencies between each label and its child labels from the sub-hierarchy sequence at once in the training step. Subsequently, two features generated from the hierarchy-aware masked self-attention and a text encoder are merged through text-hierarchy attention. Note that HiDEC does not require additional parameters for the structure encoder as used in global models, such as HiAGM, HiMatch, and HGCLR. Consequently, the parameters of the model increase linearly with respect to the classes in a hierarchy. In the inference step, sub-hierarchy decoding is recursively applied to expand from parent to child labels in a top-down manner. As a result, all parent labels at the same depth are expanded simultaneously. Thus, the maximum recursions are the depth of the entire hierarchy and not the sequence length.

A series of experiments show that HiDEC outperforms state-of-the-art (SOTA) models on two small-scale datasets, RCV1-v2 and NYT, and a large-scale dataset, EURLEX57K. Consequently, HiDEC achieves better performance with significantly fewer parameters on the three benchmark datasets. Thus, the proposed approach can solve scalability problems in large-scale hierarchies.

The contributions of this paper can be summarized as follows:
\begin{itemize}
\item This paper formulates HTC as a sub-hierarchy sequence generation using an encoder-decoder architecture. We can incorporate the hierarchy information into the sub-hierarchy sequence instead of the model structure, as all dependencies are aware by parse tree notation.
\item This paper proposes a Hierarchy DECoder (HiDEC) that recursively decodes the text sequence into a sub-hierarchy sequence by sub-hierarchy decoding while remaining aware of the path information.
\item This paper demonstrates the superiority of HiDEC by comparing SOTA models on three benchmark HTC datasets (RCV1-v2, NYT, and EURLEX57K). The results reveal the role of HiDEC in HTC through in-depth analysis.
\end{itemize}

\begin{figure*}[!t]
\centering
\includegraphics[width=1.0\textwidth]{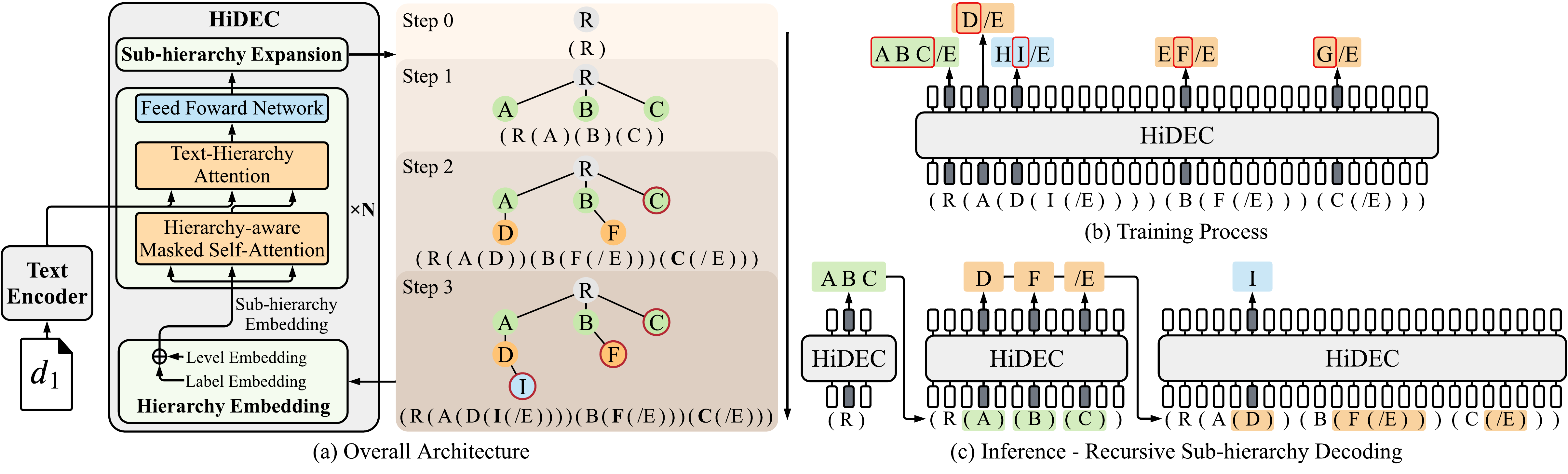} 
\caption{(a): The overall architecture of HiDEC. A feature vector from a text encoder is decoded to a sub-hierarchy sequence, which is expanded one level at once by starting from the root. (b) and (c): Illustration of input and output in training and inference, respectively. In step 3 of (a) and the right of (c), HiDEC correctly generates the sub-hierarchy (sequence) we expected. /E means “[END]” token.}
\label{fig2}
\end{figure*}

\section{Related Work}
The critical point to HTC is the use of hierarchy information, denoted as relationships among labels. For example, the relationships include root-to-target labels (path information), parent-to-child, and the entire hierarchy (holistic information). 

Research on HTC can be categorized into local and global approaches. In the local approach, a set of classifiers are used for small units of classes, such as for-each-class \cite{local_htrans}, for-each-parent \cite{local_parent,local_hdltex}, for-each-level \cite{local_hft}, and for-each-sub-hierarchy \cite{local_dgcnn}. In \cite{local_hdltex}, HDLTex was introduced as a local model that combined a deep neural network (DNN), CNN, and RNN to classify child nodes. Moreover, HTrans \cite{local_htrans} extended HDLTex to maintain path information across local classifiers based on transfer learning from parent to child. HMCN \cite{local_hmcn} applied global optimization to the classifier of each level to solve the exposure bias problem. Finally, HR-DGCNN \cite{local_dgcnn} divided the entire hierarchy into sub-hierarchies using recursive hierarchical segmentation. Unfortunately, applying this to large-scale hierarchies is challenging because many parameters are required from a set of local classifiers for small units of classes.

In the global approach, the proposed methods employed a single model with path \cite{global_capsule,global_hcsm,global_heagrcnn,global_hilap,global_hnatc,global_htcinfomax} or holistic \cite{global_hiagm,global_himatch,global_hgclr} information. For instance, HNATC \cite{global_hnatc} obtained path information using a sequence of outputs from the previous levels to predict the output at the next level. In HiLAP-RL \cite{global_hilap}, reinforcement learning was exploited in that HTC was formalized as a pathfinding problem. Moreover, HE-AGCRCNN \cite{global_heagrcnn} and HCSM \cite{global_hcsm} used capsule networks. Recent research \cite{global_hiagm,global_himatch,global_hgclr} has attempted to employ holistic information of an entire hierarchy using a structure encoder with GCN \cite{gcn} and Graphormer \cite{graphormer}. HiAGM \cite{global_hiagm} propagated text through GCN, HiMatch \cite{global_himatch} improved HiAGM by adapting semantic matching between text and label features from text features and label embeddings. In addition, HGCLR \cite{global_hgclr} attempted to unify a structure encoder with a text encoder using a novel contrastive learning method, where BERT \cite{bert} and Graphormer were employed for the text and structure encoders, respectively. Therefore, classification was performed using a hierarchy-aware text feature produced by the text encoder. Finally, it reported SOTA performance on RCV1-v2 and NYT but is infeasible in large-scale hierarchies because of the large model size caused by incorporating holistic information into the model structure.

\section{Proposed Methods}
The HTC problem can be defined using a tree structure. A hierarchy is represented as a tree $G=(V,\vec{E})$ where $V=\{v_1,v_2,… ,v_C\}$ is a set of \textit{C}-labels in the hierarchy, and $\vec{E}=\{(v_i,v_j)|v_i \in V, v_j \in child(v_i)\}$ is a set of edges between a label $v_i$ and a child $v_j$ of $v_i$. $D=\{d_1,d_2,…,d_K\}$ is a collection of $K$ documents. A document $d_k$ has a sub-hierarchy $G^{d_k}=(V^{d_k},\vec{E}^{d_k})$ converted from assigned labels where $V^{d_k}=L^{d_k}\cup\{v_i^{d_k}|v_i^{d_k} \in ancestor(v_j^{d_k}),v_j^{d_k} \in L^{d_k}\}$ and $\vec{E}^{d_{k}}=\{(v_i,v_j)|v_j \in V^{d_k}, v_i \in parent(v_j)\}$, where $L^{d_k}=\{v_1^{d_k},v_2^{d_k},\ldots,v_t^{d_k}\}$ is a label set of document $d_k$. In other words, $G^{d_k}$ is constructed using all the labels assigned to $d_k$ and their ancestors. $\hat{G}_0^{d_k}=(\{v_{root}\},\emptyset)$ is the initial sub-hierarchy of HiDEC, which has a root and no edges. Based on $\hat{G}_0^{d_k}$, recursive hierarchy decoding is defined by expanding $\hat{G}_{p}^{d_k}$ for $p$ times from $p$=0. The goal of training HiDEC is given by $\hat{G}_{p}^{d_k}=G^{d_k}$.

In Figure \ref{fig2}-(a), the overall architecture of HiDEC is presented with a demonstration of the recursive hierarchy decoding. The remainder of this section presents the details of the proposed model.

\subsection{Text Encoder}
In the proposed model, a text encoder can use any model that outputs the text feature matrix of all the input tokens, such as GRU and BERT \cite{bert}. For simplicity, let us denote $d_k$ as $\mathbf{T}=[w_1,w_2,…,w_N]$ where $w_n$ is a one-hot vector for an index of the $n$-th token. Initially, a sequence of tokens was converted into word embeddings $\mathbf{H}^0 (=\mathbf{W}^0\mathbf{T}) \in \mathbb{R}^{N\times e}$ where $\mathbf{W}^0$ is the weight matrix of the word embedding layer, and $e$ is an embedding dimension.
Given $\mathbf{H}^0$, the hidden state $\mathbf{H}$ from the text encoder can be computed using Equation \ref{eq:eq1}:
\begin{equation}
    \label{eq:eq1}
    \mathbf{H}=\text{TextEncoder}(\mathbf{H}^{0}).
\end{equation}
\subsection{Hierarchy DECoder (HiDEC)}

\subsubsection{Hierarchy Embedding Layer}
The sub-hierarchy embeddings, as shown in Figure \ref{fig1}, is obtained by initially constructing a sub-hierarchy sequence from a document $d_k$. This process consists of two steps. First, a sub-hierarchy $G^{d_k}=(V^{d_k},\vec{E}^{d_k})$ of $d_k$ is built with its target labels. Second, a sub-hierarchy sequence $\mathbf{S}$ following a parse tree notation is generated from $G^{d_k}$. Three special tokens, “(”, “)”, and “[END]”, are used to properly represent the sub-hierarchy. The tokens “(” and “)” denote the start and end of a path from each label, respectively, whereas the “[END]” token indicates the end of a path from a root. For example, $\mathbf{S}$=[( R ( A ( D ( I ( [END] ) ) ) ) ( B ( F ( [END] ) ) ) ( C ( [END] ) ) )] is constructed in Figure \ref{fig1} with a label set [C,F,I]. Once again, the tokens in $\mathbf{S}$ are represented as one-hot vectors for further processing. Subsequently, these tokens can be represented as $\mathbf{\bar{S}}=[s_1,s_2,… ,s_M]$ where $s_i=\mathbb{I}_{v}$ is a one-hot vector for a label $v$ and the special tokens. Finally, the sub-hierarchy embeddings $\mathbf{U}^0$ are constructed after explicitly incorporating the level information, similar to Transformer's position encoding \cite{transformer}, using Equations \ref{eq:eq2} and \ref{eq:eq3}:
\begin{equation}
    \label{eq:eq2}
    \mathbf{\bar{U}}^{0}=\mathbf{W}^{s}\mathbf{\bar{S}},
\end{equation}
\begin{equation}
    \label{eq:eq3}
    \mathbf{U}^{0}=\text{LevelEmbedding}(\mathbf{\bar{U}}^{0}).
\end{equation}

\subsubsection{Hierarchy-Aware Masked Self-Attention}

\begin{algorithm}[!t]
	\caption{Recursive Hierarchy Decoding in Inference} 
	\label{alg:decoding}
	\textbf{Indices:} Hierarchy depth $P$, Number of attentive layers $R$
	
	\textbf{Input:} Text feature matrix from text encoder $\mathbf{H}$
	
	\textbf{Output:} Predicted label set $L$
	\begin{algorithmic}[1]
        \Statex \textbf{//HiDEC}
	    \State $L=\emptyset$
	    \State $\hat{G}_{0}=(\{v_{root}\},\emptyset)$
        \For{$p=0,\ldots,P-1$}
            \Statex \textbf{//Sub-hierarchy embedding}
            \State Convert $\hat{G}_{p}$ to sub-hierarchy sequence $\mathbf{S}_{p}$
            \State Compute $\mathbf{U}^{0}$ from $\mathbf{S}_{p}$ with Eq.\ref{eq:eq2}, \ref{eq:eq3}
            \State Generate masking matrix $\mathbf{M}$ with Eq.\ref{eq:eq5}
            \Statex \textbf{//Attentive layers}
            \For{$r=0,\ldots,R-1$}
                \State $\mathbf{U}^{r+1}$ = Attention($\mathbf{U}^r,\mathbf{H},\mathbf{M}$) with Eq.\ref{eq:eq4}, \ref{eq:eq6}, \ref{eq:eq7}
            \EndFor
            \State $\mathbf{U}=\mathbf{U}^R$
            \Statex \textbf{//Sub-hierarchy expansion}
            \For{$s_i \in S_p$}
                \If{$s_i \notin $ special token set}
                    \State $v_i=s_i$
                    \For{$v_j \in child(v_i)$}
                        \State $c_{ij}=U_i\cdot \mathbf{W}^S \cdot \mathbb{I}_{v_{j}}$ with Eq.\ref{eq:eq8}
                        \State $p_i=\text{sigmoid}(c_i)$ with Eq.\ref{eq:eq9}
                        \State Get $\hat{y}_i$ from $p_i$ by thresholding
                        \For{$v_k \in \hat{y}_i$}
                            \State $V_{p}^{}=V_{p}^{}\cup\{v_k\}$
                            \State $\vec{E}_p=\vec{E}_p\cup\{(v_i, v_k)\}$
                        \EndFor
                    \EndFor
                \EndIf
            \EndFor
            \State $\hat{G}_{p+1}=(V_p,\vec{E}_p)$
        \EndFor
        \Statex \textbf{//Label assignment}
        \For{$i=0,\ldots,|V_P|$}
            \If{$v_i\in {leaf}(\hat{G}_P)$}
                \State $L=L\cup\{v_i\}$
            \Else{\textbf{ if} $v_i==${“[END]”} \textbf{then}}
                \State $L=L\cup\{parent(v_i)\}$
            \EndIf
        \EndFor
        \State \textbf{return} $L$
	\end{algorithmic}
\end{algorithm}

This component is responsible for capturing hierarchy information, similar to the structure encoder in global models \cite{global_hiagm, global_himatch, global_hgclr}. However, only a sub-hierarchy from the entire hierarchy, which is thought to be highly relevant information for classification, is used based on the self-attention mechanism used by Transformer \cite{transformer}. To compute self-attention scores, we applied hierarchy-aware masking to incorporate hierarchy information. The self-attention mechanism of Transformer was exploited with a minor modification concerning hierarchy-aware masking. The hierarchy-aware masked self-attention of $r$-th layer is computed using Equation \ref{eq:eq4}:
\begin{equation}
    \label{eq:eq4}
    \Dot{\mathbf{U}}^{r}=\text{MHA}(\mathbf{W}_{Q}^{r}\mathbf{U}^{r-1},\mathbf{W}_{K}^{r}\mathbf{U}^{r-1},\mathbf{W}_{V}^{r}\mathbf{U}^{r-1},\mathbf{M}),
\end{equation}
where MHA is the multi-head attention, the same as that of Transformer. $\mathbf{W}_Q^{r}$,$\mathbf{W}_K^{r}$,$\mathbf{W}_V^{r}$ are projection weight matrices for the query, key, and value, respectively. Moreover, $\mathbf{M}$ is the hierarchy-aware mask defined as follows:
\begin{equation}
    \label{eq:eq5}
    \mathbf{M}_{ij}=\begin{cases}
    -1e9 & \text{ if } v_i \notin ancestor(v_j) \\ 
    0 & \text{ else }
    \end{cases}.
\end{equation}

We ignore the dependency between two labels if they are not the same label and not an ancestor by setting $\mathbf{M}_{ij}=-1e9$. This setting makes the model attend to the path information relevant to the input documents and ignores the hierarchy information at the lower-level labels than each label to learn a sub-hierarchy sequence at once in a training step. Note that the dependencies of the three special tokens with respect to the other tokens, including themselves, are considered.

\subsubsection{Text-Hierarchy Attention}
In text-hierarchy attention, we can compute the attention scores of labels by dynamically reflecting the importance of tokens in an input document. A new sub-hierarchy matrix $\ddot{\mathbf{U}}^r$ of $r$-th layer is computed by combining the text feature matrix $\mathbf{H}$ from the encoder and $\dot{\mathbf{U}}^{r}$ without a masking mechanism using Equation \ref{eq:eq6}:
\begin{equation}
    \label{eq:eq6}
    \ddot{\mathbf{U}}^{r}=\text{MHA}(\mathbf{W}_{Q}^{r}\dot{\mathbf{U}}^{r},\mathbf{W}_{K}^{r}\mathbf{H},\mathbf{W}_{V}^{r}\mathbf{H},-).
\end{equation}

Subsequently, the output of $r$-th layer $\mathbf{U}^r$ is obtained using a position-wise feed-forward network (FFN) using Equation \ref{eq:eq7}:
\begin{equation}
    \label{eq:eq7}
    \mathbf{U}^{r}=\text{FFN}(\ddot{\mathbf{U}}^{r}).
\end{equation}

Consequently, the output of the final layer, $\mathbf{U}$, in HiDEC is used in the sub-hierarchy expansion.

\subsubsection{Sub-Hierarchy Decoding}

Sub-hierarchy decoding is crucial in generating a sub-hierarchy using recursive hierarchy decoding. This results in a target sub-hierarchy if HiDEC functions as expected. For each label, the classification to child labels is performed using the sub-hierarchy matrix $\mathbf{U}$ using Equations \ref{eq:eq8} and \ref{eq:eq9}:
\begin{equation}
    \label{eq:eq8}
    \begin{matrix}
    c_{ij}=U_i\cdot \mathbf{W}^S \cdot \mathbb{I}_{v_j} & \forall v_j \in child(v_i)
    \end{matrix},
\end{equation}
\begin{equation}
    \label{eq:eq9}
    p_i=\text{sigmoid}(c_i),
\end{equation}
where $c_{ij}$ is a similarity score of child $v_j$ under a parent $v_i$, and $p_i$ is the probability of a child $v_j$ obtained using a task-specific probability function such as sigmoid. The three special tokens are excluded when selecting the parent $v_i$.

We reduced the label space of HTC by focusing on the child labels of a parent label of interest. During training, we use binary cross-entropy loss functions as shown in Equation \ref{eq:eq10}:
\begin{equation}
    \label{eq:eq10}
    \mathcal{L}=-\frac{1}{MJ}\sum_{i=0}^{M}\sum_{j=0}^{J}y_{ij}\text{log}(p_{ij})+(1-y_{ij})\text{log}(1-p_{ij}),
\end{equation}
where $J=|child(v_i)|$ indicates the number of child labels for parent $v_i$. Moreover, $y_{ij}$ and $p_{ij}$ denote a target label of $j$-th child label of $v_i$ and its output probability, respectively.

At the inference time, recursive hierarchy decoding is performed using a threshold. The details of the recursive hierarchy decoding are described in Algorithm \ref{alg:decoding}. The number of decoding steps is the same as the maximum depth of the hierarchy. At each decoding step, all tokens except the special tokens are expanded. Decoding ends if the tokens are leaf labels or “[END]”. Finally, the labels associated with “[END]” or leaf labels are assigned to the input as predictions.
\section{Experiments}

\begingroup
\setlength{\tabcolsep}{2.4pt}
\begin{table}[!t]
\begin{tabular}{crccrrr}
\toprule
{Dataset} & \multicolumn{1}{c}{{$|\textit{L}|$}} & {Depth} & {Avg} & \multicolumn{1}{c}{{Train}} & \multicolumn{1}{c}{{Val}} & \multicolumn{1}{c}{{Test}} \\ \toprule
RCV1-v2     &103          &4              &3.24                                 &20,833          &2,316         &781,265        \\
NYT     &166          &8              &7.60                                    &23,345          &5,834         &7,292          \\
EURLEX57K     &4,271         &5              &5.00                                 &45,000         &6,000         &6,000         \\ 
\bottomrule
\end{tabular}
\caption{Data statistics. $|\textit{L}|$ denotes the number of labels. Depth and Avg are the maximum hierarchy depth and the average number of assigned labels for each text, respectively.}
\label{tab:table1}
\end{table}
\endgroup

\subsection{Datasets and Evaluation Metrics}
Table \ref{tab:table1} lists the data statistics used in the experiments. For the standard evaluation, two small-scale datasets, RCV1-v2 \cite{rcv} and NYT \cite{nyt}, and one large-scale dataset, EURLEX57K \cite{eurlex}, were chosen. RCV1-v2 comprises 804,414 news documents, divided into 23,149 and 781,265 documents for training and testing, respectively, as benchmark splits. We randomly sampled 10\% of the training data as the validation data for model selection. NYT comprises 36,471 news documents divided into 29,179 and 7,292 documents for training and testing, respectively. For a fair comparison, we followed the data configurations of previous work \cite{global_hiagm, global_himatch}. In particular, EURLEX57K is a large-scale hierarchy with 57,000 documents and 4,271 labels. Benchmark splits of 45,000, 6,000, and 6,000 were used for training, validation, and testing, respectively. We used Micro-F1 for three datasets and Macro-F1 for RCV1-v2 and NYT.

\begingroup
\setlength{\tabcolsep}{1pt}
\begin{table*}[!t]
    \centering
    \begin{minipage}{1\linewidth}
        \begin{minipage}{.53\linewidth}
            \centering
                \begin{tabular}{ccccc}
                    \toprule
                    \multirow{2}{*}{{Model}} & \multicolumn{2}{c}{{RCV1-v2}} & \multicolumn{2}{c}{{NYT}} \\ 
                    \cline{2-5} & {Micro} & {Macro} & {Micro} & {Macro} \\ \toprule
                    \multicolumn{5}{c}{{w/o Pretrained Language Models}}  \\ \toprule
                    TextRCNN$^*$ \cite{global_hiagm} &81.57  &59.25  &70.83  &56.18\\
                    HiAGM \cite{global_hiagm}	&83.96	&63.35	&74.97	&60.83   \\
                    HTCInfoMax \cite{global_htcinfomax}	&83.51	&62.71	&-	&-   \\
                    HiMatch \cite{global_himatch}	&84.73	&64.11	&-	&-   \\
                    HiDEC	&\textbf{85.54}	&\textbf{65.08}	&\textbf{76.42}	&\textbf{63.99}   \\
                    \toprule
                    \multicolumn{5}{c}{{w/ Pretrained Language Models}}  \\ \toprule
                    BERT$^*$ \cite{global_hgclr} &85.65	&67.02	&78.24	&65.62   \\
                    HiAGM \cite{global_hgclr}	&85.58	&67.93	&78.64	&66.76   \\
                    HTCInfoMax \cite{global_hgclr}	&85.53	&67.09	&78.75	&67.31   \\
                    HiMatch \cite{global_himatch}	&86.33	&68.66	&-	&-   \\
                    HGCLR \cite{global_hgclr}	&86.49	&68.31	&78.86	&67.96   \\
                    HiDEC	&\textbf{87.96}	&\textbf{69.97}	&\textbf{79.99}	&\textbf{69.64}   \\
                    \bottomrule
                    \multicolumn{5}{c}{(a)}
                \end{tabular}
        \end{minipage}%
        \begin{minipage}{0.47\linewidth}
            \centering
                \begin{tabular}{cc}
                    \toprule
                    \multirow{2}{*}{{Model}} & {EURLEX57K}\\ 
                    \cline{2-2} & {Micro}\\ \toprule
                    \multicolumn{2}{c}{{w/o Pretrained Language Models}}  \\ \toprule
                    BiGRU-ATT$^*$ \cite{eurlex}	&68.90\\
                    HAN$^*$ \cite{eurlex}	&68.00\\
                    CNN-LWAN$^*$ \cite{eurlex}	&64.20\\
                    BiGRU-LWAN$^*$ \cite{eurlex}	&69.80\\
                    HiMatch	&71.11\\
                    HiDEC	&\textbf{71.23}\\
                    \toprule
                    \multicolumn{2}{c}{{w/ Pretrained Language Models}}  \\ \toprule
                    BERT$^*$	&73.20\\
                    HiDEC	&\textbf{75.29}\\
                    \bottomrule
                    \multicolumn{2}{c}{(b)}
                \end{tabular}
        \end{minipage}
    \end{minipage}
    \caption{Performance comparison on three datasets, (a) RCV1-v2 and NYT, (b) EURLEX57K. $^*$ denotes models without hierarchy information.}
\label{tab:table2}
\end{table*}
\endgroup

\subsection{Implementation Details}
After text cleaning and stopword removal, words with two or more occurrences were selected to retain the vocabulary. Consequently, the vocabulary sizes for the three benchmark datasets were 60,000.

For the text encoder, we opted the simplest encoder, a bi-directional GRU with a single layer, for a fair comparison to previous work \cite{global_hiagm, global_himatch}. The size of the hidden state was set to 300. The word embeddings in the text encoder were initialized using 300-dimensional GloVe \cite{glove}. In contrast to the GRU-based encoder, we used BERT, bert-base-uncased \cite{bert}, to demonstrate the generalization ability of the pre-trained encoder. The output hidden matrix from the last layer of BERT was used as the context matrix $\mathbf{H}$ in Equation \ref{eq:eq1}.

For HiDEC, a layer with two heads were used for both GRU-based encoder and BERT. The label and level embeddings with 300- and 768-dimension for the GRU-based encoder and BERT, respectively, were initialized using a normal distribution with $\mu$=0 and $\sigma$=$300^{-0.5}$. The hidden state size in the attentive layer was the same as the label embedding size. The FFN comprised two FC layers with 600- and 3,072-dimension feed-forward filter for the GRU-based encoder and BERT, respectively. Based on an empirical test, we removed the residual connection from the original Transformer decoder \cite{transformer}. The threshold for recursive hierarchy decoding was set to 0.5. A dropout with a probability of 0.5, 0.1, and 0.1 was applied to the embedding layer and behind every FFN and attention, respectively.

For optimization, Adam optimizer \cite{adam} was utilized with learning rate $lr$=1e-4, $\beta_1$=0.9, $\beta_2$=0.999, and $eps$=1e-8. The size of the mini-batch was set to 256 for GRU-based models. With BERT as a text encoder model, we set $lr$ and the mini-batch size to 5e-5 and 64, respectively. The $lr$ was controlled using a linear schedule with a warm-up rate of 0.1. Gradient clipping with a maximum gradient norm of 1.0 was performed to prevent gradient overflow.

All models were implemented using PyTorch \cite{pytorch} and trained using NVIDIA A6000. The average score of the ten different models was utilized as the proposed model performance, where the model with the best performance was selected using Micro-F1 on the validation data. 

\begingroup
\begin{table}[]
\centering
            \begin{tabular}{cccr}
                \toprule
                \multirow{2}{*}{{Model}} &{RCV1-v2} &{NYT} &\multicolumn{1}{c}{{EURLEX57K}} \\ 
                \cline{2-4} &103 &166 &\multicolumn{1}{c}{4,271} \\ \toprule
                \multicolumn{4}{c}{{w/o Pretrained Language Models}}  \\ \toprule
                TextRCNN$^*$	&18M	&18M	&19M    \\
                HiAGM	&31M	&42M	&5,915M  \\
                HiMatch	&37M	&52M	&6,211M  \\
                HiDEC	&\textbf{20M}	&\textbf{20M}	&\textbf{21M}    \\
                \toprule
                \multicolumn{4}{c}{{w/ Pretrained Language Models}}  \\ \toprule
                BERT$^*$	&109M	&109M	&112M   \\
                HGCLR	&\textbf{120M}	&\textbf{121M}	& 425M   \\
                HiDEC	&123M	&123M	&\textbf{127M}   \\
                \bottomrule
            \end{tabular}
            \caption{Model parameter comparison. $^*$ denotes models without hierarchy information.}
\label{tab:table3}
\end{table}
\endgroup

\subsection{Comparison Models}
We selected various baseline models from recent work. For RCV1-v2 and NYT, TextRCNN \cite{textrcnn}, HiAGM \cite{global_hiagm}, HiMatch \cite{global_himatch}, HTCInfoMax \cite{global_htcinfomax}, and HGCLR \cite{global_hgclr} were chosen. TextRCNN comprises bi-GRU and CNN layers and is a hierarchy-unaware model used as a text encoder in HiAGM and HiMatch. HiAGM combines text and label features from the text encoder and  GCN-based structure encoder, respectively, using text propagation. HTCInfoMax and HiMatch improved HiAGM with a prior distribution and semantic matching loss, respectively. They can be improved by replacing the text encoder with PLM. HGCLR directly embeds hierarchy information into the text encoder using Graphormer during training. For EURLEX57K, BiGRU-ATT \cite{bigruatt}, HAN \cite{han}, CNN-LWAN \cite{cnn-lwan}, BiGRU-LWAN \cite{eurlex}, and HiMatch \cite{global_himatch} were chosen. BiGRU-ATT and HAN are strong baselines for text classification tasks with an attention mechanism. CNN and BiGRU-LWAN extended BiGRU-ATT with label-wise attention. Owing to the large model size, applying HiMatch to EURLEX57K is infeasible. According to the paper \cite{global_himatch}, the text-propagation module weakly influences the performance compared with the original approach. Therefore, we simplified HiMatch by removing the text-propagation module.

\subsection{Experimental Results}
Table \ref{fig2} summarizes the performance of HiDEC and the other models on (a) two small-scale datasets, RCV1-v2 and NYT, and (b) a large-scale dataset, EURLEX57K. HiDEC achieved the best performance on the three datasets regardless of whether BERT was used \cite{bert} as the text encoder. This highlights the effectiveness of HiDEC using sub-hierarchy information rather than entire-hierarchy information. In addition, we found that HiDEC highly benefits from PLM compared to other models, as the performance gains of HiDEC with and without BERT are relatively large. For example, in MicroF1 on RCV1-v2, the gain between HiDEC and HiDEC with PLM was 2.42. In contrast, the gain between HiMatch \cite{global_himatch} and HiMatch with PLM was 1.60. On EURLEX57K, training the hierarchy-aware models was infeasible because of the large model size\footnote{Except for HiDEC and simplified HiMatch, we encountered out-of-memory with NVIDIA A6000 48GB for other models in training.} caused by the structure encoder for 4,271 labels, except for HiDEC and simplified HiMatch. However, this does not apply to BERT because of the model size. Similar to RCV1-v2 and NYT, HiDEC with BERT exhibited the best performance.

\begingroup
\setlength{\tabcolsep}{1.9pt}
\begin{table}[]
    \centering
    \begin{tabular}{cccc}
        \toprule
        \multirow{2}{*}{{Model}} & \multicolumn{2}{c}{{RCV1-v2}} & {EURLEX57K} \\ 
        \cline{2-4} & {Micro} & {Macro} & {Micro} \\ \toprule
        \multicolumn{1}{l}{HiDEC}	&\textbf{85.54}	&\textbf{64.04}	&\textbf{71.31}  \\
        \multicolumn{1}{l}{\;$+$ Residual connection}	&84.69	&60.32	&69.97  \\
        \multicolumn{1}{l}{\;$-$ Hierarchy-aware masking}	&85.46	&63.73	&71.22  \\
        \multicolumn{1}{l}{\;$-$ Level Embedding}	&85.51	&63.91	&70.90  \\
        \bottomrule
    \end{tabular}
\caption{Ablation studies on RCV1-v2 and EURLEX57K.}
\label{tab:table4}
\end{table}
\endgroup

\begin{figure*}[!t]
\centering
\includegraphics[width=0.97\textwidth]{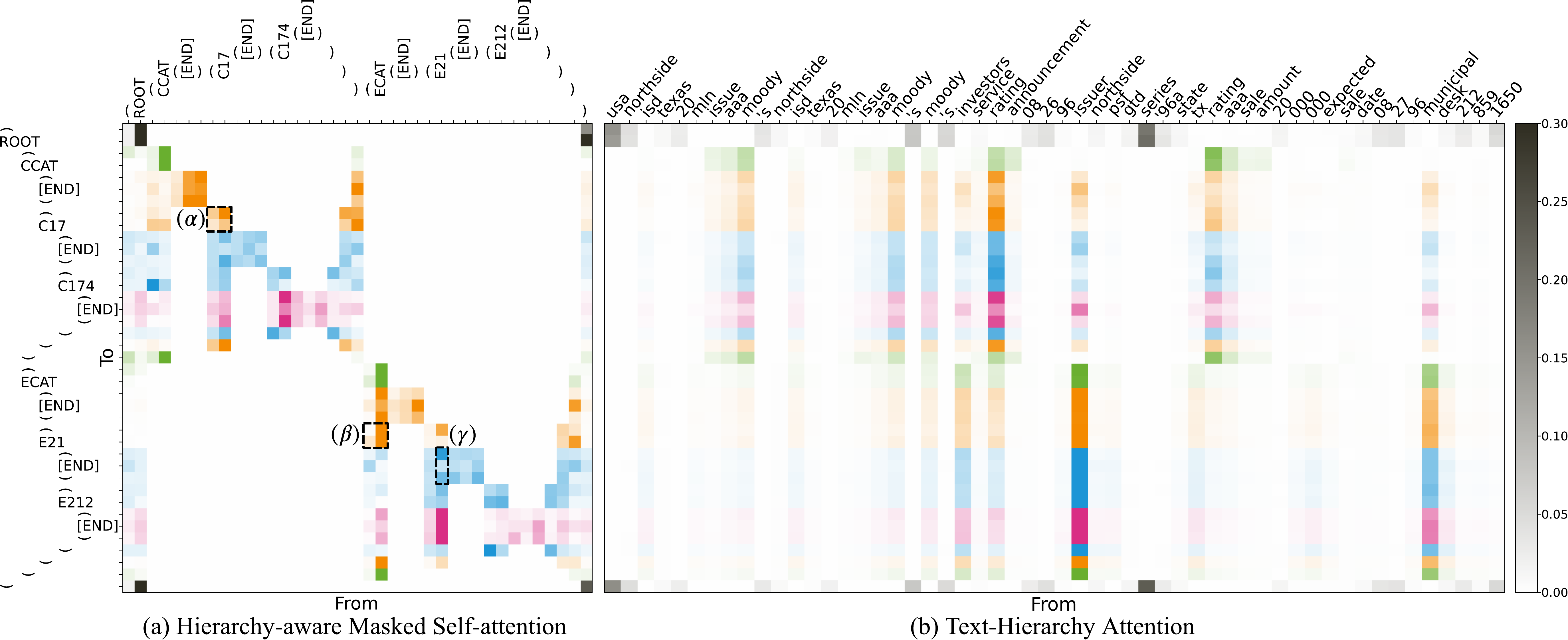} 
\caption{Heatmaps of the attention scores in HiDEC on RCV1-v2. (a) and (b) are heatmap of the hierarchy-aware masked self-attention scores and text-hierarchy attention scores, respectively. Attention scores over 0.3 are clipped in all the heatmaps and the similar colors indicate the same level.}
\label{fig3}
\end{figure*}

\begin{figure}[!t]
\centering
\includegraphics[width=0.85\columnwidth]{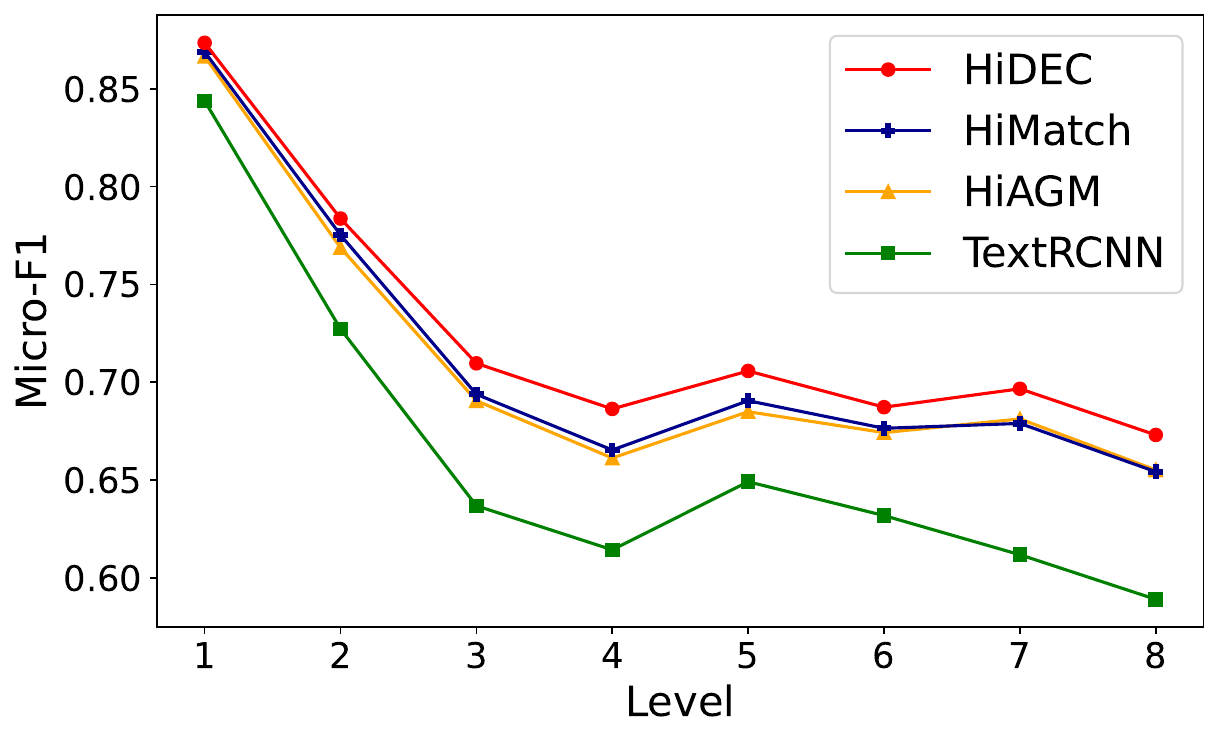} 
\caption{The level-wise performance of GRU-based HTC models on NYT.}
\label{fig4}
\end{figure}

\subsection{Model Parameters}
Table \ref{tab:table3} summarizes the parameters for different models on three benchmark datasets. The label sizes are RCV1-v2 (103) \texttt{<} NYT (166) \texttt{<<} EURLEX57K (4,271). The table shows that the parameters of the existing models increase dramatically as the label size increases. Note that HiDEC requires significantly fewer parameters even though the label size increases. In the extreme case on EURLEX57K, HiDEC only requires 21M parameters, which are 295x smaller than HiMatch with 6,211M. Consequently, the parameters in HiDEC increase linearly with respect to the labels in a hierarchy because no extra parameters are required for the structure encoder and sub-tasks. HiAGM and HiMatch require merging parameters projecting text features into label features for text propagation. In addition, HGCLR needs edge and spatial encoding parameters for the structure encoder. However, HiDEC does not require these parameters because attentive layers play the same role. Only the label embeddings increase according to the label size in a hierarchy.

\subsection{Ablation Studies}
Table \ref{tab:table4} shows the ablation studies of each component in HiDEC without PLM on RCV1-v2 and EURLEX57K. HiDEC differs from the original Transformer decoder \cite{transformer} in the absence of a residual connection and the existence of hierarchy-aware masking and level embedding. We observed that adding the residual connection and eliminating hierarchy-aware masking and level embedding also had a negative effect. Among them, adding a residual connection had the most negative effect. We presumed that the essential information from previous features for HTC was hindered by the residual connection.

\subsection{Interpretation of Attentions}
We investigated the roles of hierarchy-aware masked self-attention and text-hierarchy attention by visualizing the attention scores on RCV1-v2, as shown in Figure \ref{fig3}. The self-attention scores are shown in Figure \ref{fig3}-(a). In ($\alpha$), the attention score between “(” and “C17” is relatively high where “(” is a starting path from “C17”. The score shows that the special tokens “(” and “)” were appropriately associated with the corresponding labels. In ($\beta$), the dependency between child “E21” and parent “ECAT” was well-described because the attention scores for child labels under parent “ECAT” were high. In ($\gamma$), the results show a dependency between the label assignment sequence – [“(”, “END”, “)”] and the label “E21”. From these three examples, we can conclude that hierarchy-aware masked self-attention effectively captures the path dependencies. Figure \ref{fig3}-(b) shows the attention scores between the input tokens and a sub-hierarchy sequence. Some tokens, such as the “rating” and “moody,” have high attention scores for the descendants of “CCAT” and itself, where “CCAT” denotes “CORPORATE/INDUSTRIAL”. In contrast, some tokens like “issuer,” “municipal,” and “investors” have high attention scores for the descendants of “EACT” and itself, where “EACT” denotes “ECONOMICS”. This result indicates that the labels are associated with different tokens to different degrees.
 
\subsection{Level-Wise Performance}
Figure \ref{fig4} depicts the level-wise performance of the models using a GRU-based text encoder on NYT. It shows the effectiveness of the hierarchy-aware models by comparing TextRCNN increases as the level increases. Among them, HiDEC consistently achieved the best performance at all levels. Note that significant improvements were obtained at the deep levels, implying that sub-hierarchy information is more powerful in capturing the structure information of a target document than the entire hierarchy information. 

\section{Conclusion}
This paper addressed the scalability limitations of recent HTC models due to the large model size of the structure encoders. To solve this problem, we formulated HTC as a sub-hierarchy sequence generation using an encoder-decoder architecture. Subsequently, we propose Hierarchy DECoder (HiDEC) which recursively decodes the text sequence into a sub-hierarchy sequence by sub-hierarchy decoding while staying aware of the path information. HiDEC achieved state-of-the-art performance with significantly fewer model parameters than existing models on benchmark datasets, such as RCV1-v2, NYT, and EURLEX57K. In the future, we plan to extend the proposed model to extremely large-scale hierarchies (e.g., MeSH term indexing or product navigation) and introduce a novel training strategy combining top-down and bottom-up methods that can effectively use a hierarchy structure.

\section{Acknowledgments}
This work was supported by the National Research Foundation of Korea (NRF) grant funded by the Korea government(MSIT) (No. NRF-2019R1G1A1003312) and (No. NRF-2021R1I1A3052815).

\bibliography{HiDEC}

\begin{thebibliography}{33}
\providecommand{\natexlab}[1]{#1}

\bibitem[{Banerjee et~al.(2019)Banerjee, Akkaya, Perez-Sorrosal, and
  Tsioutsiouliklis}]{local_htrans}
Banerjee, S.; Akkaya, C.; Perez-Sorrosal, F.; and Tsioutsiouliklis, K. 2019.
\newblock Hierarchical Transfer Learning for Multi-label Text Classification.
\newblock In \emph{Proceedings of the 57th Annual Meeting of the Association
  for Computational Linguistics}, 6295--6300. Association for Computational
  Linguistics.

\bibitem[{Cevahir and Murakami(2016)}]{product2}
Cevahir, A.; and Murakami, K. 2016.
\newblock Large-scale Multi-class and Hierarchical Product Categorization for
  an E-commerce Giant.
\newblock In \emph{Proceedings of COLING 2016, the 26th International
  Conference on Computational Linguistics: Technical Papers}, 525--535. The
  COLING 2016 Organizing Committee.

\bibitem[{Chalkidis et~al.(2019)Chalkidis, Fergadiotis, Malakasiotis, and
  Androutsopoulos}]{eurlex}
Chalkidis, I.; Fergadiotis, E.; Malakasiotis, P.; and Androutsopoulos, I. 2019.
\newblock Large-Scale Multi-Label Text Classification on EU Legislation.
\newblock In \emph{Proceedings of the 57th Annual Meeting of the Association
  for Computational Linguistics}, 6314--6322. Association for Computational
  Linguistics.

\bibitem[{Chen et~al.(2021)Chen, Ma, Lin, and Yan}]{global_himatch}
Chen, H.; Ma, Q.; Lin, Z.; and Yan, J. 2021.
\newblock Hierarchy-aware Label Semantics Matching Network for Hierarchical
  Text Classification.
\newblock In \emph{Proceedings of the 59th Annual Meeting of the Association
  for Computational Linguistics and the 11th International Joint Conference on
  Natural Language Processing (Volume 1: Long Papers)}, 4370--4379. Association
  for Computational Linguistics.

\bibitem[{Deng et~al.(2021)Deng, Peng, He, Li, and Yu}]{global_htcinfomax}
Deng, Z.; Peng, H.; He, D.; Li, J.; and Yu, P. 2021.
\newblock HTCInfoMax: A Global Model for Hierarchical Text Classification via
  Information Maximization.
\newblock In \emph{Proceedings of the 2021 Conference of the North American
  Chapter of the Association for Computational Linguistics: Human Language
  Technologies}, 3259--3265. Association for Computational Linguistics.

\bibitem[{Devlin et~al.(2019)Devlin, Chang, Lee, and Toutanova}]{bert}
Devlin, J.; Chang, M.-W.; Lee, K.; and Toutanova, K. 2019.
\newblock BERT: Pre-training of Deep Bidirectional Transformers for Language
  Understanding.
\newblock In \emph{Proceedings of the 2019 Conference of the North American
  Chapter of the Association for Computational Linguistics: Human Language
  Technologies, Volume 1 (Long and Short Papers)}, 4171--4186. Association for
  Computational Linguistics.

\bibitem[{Dumais and Chen(2000)}]{local_parent}
Dumais, S.; and Chen, H. 2000.
\newblock Hierarchical Classification of Web Content.
\newblock In \emph{Proceedings of the 23rd Annual International ACM SIGIR
  Conference on Research and Development in Information Retrieval}, 256--263.
  Association for Computing Machinery.
\newblock ISBN 1581132263.

\bibitem[{Kingma and Ba(2015)}]{adam}
Kingma, D.~P.; and Ba, J.~L. 2015.
\newblock Adam: A method for stochastic optimization.
\newblock In \emph{3rd International Conference on Learning Representations,
  ICLR 2015 - Conference Track Proceedings}. International Conference on
  Learning Representations, ICLR.

\bibitem[{Kipf and Welling(2017)}]{gcn}
Kipf, T.~N.; and Welling, M. 2017.
\newblock Semi-Supervised Classification with Graph Convolutional Networks.
\newblock In \emph{5th International Conference on Learning Representations,
  ICLR 2017, Toulon, France, April 24-26, 2017, Conference Track Proceedings}.
  OpenReview.net.

\bibitem[{Kowsari et~al.(2017)Kowsari, Brown, Heidarysafa, Meimandi, Gerber,
  and Barnes}]{local_hdltex}
Kowsari, K.; Brown, D.~E.; Heidarysafa, M.; Meimandi, K.~J.; Gerber, M.~S.; and
  Barnes, L.~E. 2017.
\newblock HDLTex: Hierarchical Deep Learning for Text Classification.
\newblock In \emph{2017 16th IEEE International Conference on Machine Learning
  and Applications (ICMLA)}, 364--371.

\bibitem[{Kozareva(2015)}]{product1}
Kozareva, Z. 2015.
\newblock Everyone Likes Shopping! Multi-class Product Categorization for
  e-Commerce.
\newblock In \emph{Proceedings of the 2015 Conference of the North American
  Chapter of the Association for Computational Linguistics: Human Language
  Technologies}, 1329--1333. Association for Computational Linguistics.

\bibitem[{Lai et~al.(2015)Lai, Xu, Liu, and Zhao}]{textrcnn}
Lai, S.; Xu, L.; Liu, K.; and Zhao, J. 2015.
\newblock Recurrent Convolutional Neural Networks for Text Classification.
\newblock In \emph{Proceedings of the Twenty-Ninth AAAI Conference on
  Artificial Intelligence}, 2267--2273. AAAI Press.
\newblock ISBN 0262511290.

\bibitem[{Lewis et~al.(2004)Lewis, Yang, Rose, Li, and LEWIS}]{rcv}
Lewis, D.~D.; Yang, Y.; Rose, T.~G.; Li, F.; and LEWIS, F.~L. 2004.
\newblock RCV1: A New Benchmark Collection for Text Categorization Research.
\newblock \emph{journal of Machine Learning Research}, 5: 361--397.

\bibitem[{Mao et~al.(2019)Mao, Tian, Han, and Ren}]{global_hilap}
Mao, Y.; Tian, J.; Han, J.; and Ren, X. 2019.
\newblock Hierarchical Text Classification with Reinforced Label Assignment.
\newblock In \emph{Proceedings of the 2019 Conference on Empirical Methods in
  Natural Language Processing and the 9th International Joint Conference on
  Natural Language Processing (EMNLP-IJCNLP)}, 445--455. Association for
  Computational Linguistics.

\bibitem[{Mullenbach et~al.(2018)Mullenbach, Wiegreffe, Duke, Sun, and
  Eisenstein}]{cnn-lwan}
Mullenbach, J.; Wiegreffe, S.; Duke, J.; Sun, J.; and Eisenstein, J. 2018.
\newblock Explainable Prediction of Medical Codes from Clinical Text.
\newblock In \emph{Proceedings of the 2018 Conference of the North American
  Chapter of the Association for Computational Linguistics: Human Language
  Technologies, Volume 1 (Long Papers)}, 1101--1111. Association for
  Computational Linguistics.

\bibitem[{Paszke et~al.(2019)Paszke, Gross, Massa, Lerer, Bradbury, Chanan,
  Killeen, Lin, Gimelshein, Antiga, Desmaison, Kopf, Yang, DeVito, Raison,
  Tejani, Chilamkurthy, Steiner, Fang, Bai, and Chintala}]{pytorch}
Paszke, A.; Gross, S.; Massa, F.; Lerer, A.; Bradbury, J.; Chanan, G.; Killeen,
  T.; Lin, Z.; Gimelshein, N.; Antiga, L.; Desmaison, A.; Kopf, A.; Yang, E.;
  DeVito, Z.; Raison, M.; Tejani, A.; Chilamkurthy, S.; Steiner, B.; Fang, L.;
  Bai, J.; and Chintala, S. 2019.
\newblock PyTorch: An Imperative Style, High-Performance Deep Learning Library.
\newblock In Wallach, H.; Larochelle, H.; Beygelzimer, A.; d~Alché-Buc, F.;
  Fox, E.; and Garnett, R., eds., \emph{Advances in Neural Information
  Processing Systems 32}, 8024--8035. Curran Associates, Inc.

\bibitem[{Peng et~al.(2018)Peng, Li, He, Liu, Bao, Wang, Song, and
  Yang}]{local_dgcnn}
Peng, H.; Li, J.; He, Y.; Liu, Y.; Bao, M.; Wang, L.; Song, Y.; and Yang, Q.
  2018.
\newblock Large-Scale Hierarchical Text Classification with Recursively
  Regularized Deep Graph-CNN.
\newblock In \emph{Proceedings of the 2018 World Wide Web Conference},
  1063--1072. International World Wide Web Conferences Steering Committee.
\newblock ISBN 9781450356398.

\bibitem[{Peng et~al.(2021)Peng, Li, Wang, Wang, Gong, Yang, Li, Yu, and
  He}]{global_heagrcnn}
Peng, H.; Li, J.; Wang, S.; Wang, L.; Gong, Q.; Yang, R.; Li, B.; Yu, P.~S.;
  and He, L. 2021.
\newblock Hierarchical Taxonomy-Aware and Attentional Graph Capsule RCNNs for
  Large-Scale Multi-Label Text Classification.
\newblock \emph{IEEE Transactions on Knowledge and Data Engineering}, 33:
  2505--2519.

\bibitem[{Pennington, Socher, and Manning(2014)}]{glove}
Pennington, J.; Socher, R.; and Manning, C. 2014.
\newblock GloVe: Global Vectors for Word Representation.
\newblock In \emph{Proceedings of the 2014 Conference on Empirical Methods in
  Natural Language Processing (EMNLP)}, 1532--1543. Association for
  Computational Linguistics.

\bibitem[{Sandhaus.(2008)}]{nyt}
Sandhaus., E. 2008.
\newblock The new york times annotated corpus LDC2008T19. Web Download.
\newblock Linguistic Data Consortium.

\bibitem[{Shimura, Li, and Fukumoto(2018)}]{local_hft}
Shimura, K.; Li, J.; and Fukumoto, F. 2018.
\newblock HFT-CNN: Learning Hierarchical Category Structure for Multi-label
  Short Text Categorization.
\newblock In \emph{Proceedings of the 2018 Conference on Empirical Methods in
  Natural Language Processing}, 811--816. Association for Computational
  Linguistics.

\bibitem[{Sinha et~al.(2018)Sinha, Dong, Cheung, and Ruths}]{global_hnatc}
Sinha, K.; Dong, Y.; Cheung, J.~C.; and Ruths, D. 2018.
\newblock A Hierarchical Neural Attention-based Text Classifier.
\newblock In \emph{Proceedings of the 2018 Conference on Empirical Methods in
  Natural Language Processing, EMNLP 2018}, 817--823. Association for
  Computational Linguistics.
\newblock ISBN 9781948087841.

\bibitem[{Vaswani et~al.(2017)Vaswani, Shazeer, Parmar, Uszkoreit, Jones,
  Gomez, Kaiser, and Polosukhin}]{transformer}
Vaswani, A.; Shazeer, N.; Parmar, N.; Uszkoreit, J.; Jones, L.; Gomez, A.~N.;
  Kaiser, L.; and Polosukhin, I. 2017.
\newblock Attention is All you Need.
\newblock In Guyon, I.; Luxburg, U.~V.; Bengio, S.; Wallach, H.; Fergus, R.;
  Vishwanathan, S.; and Garnett, R., eds., \emph{Advances in Neural Information
  Processing Systems}, volume~30. Curran Associates, Inc.

\bibitem[{Wang et~al.(2021)Wang, Hu, Li, and Yu}]{global_hcsm}
Wang, B.; Hu, X.; Li, P.; and Yu, P.~S. 2021.
\newblock Cognitive structure learning model for hierarchical multi-label text
  classification.
\newblock \emph{Knowledge-Based Systems}, 218: 106876.

\bibitem[{Wang et~al.(2022)Wang, Wang, Huang, Sun, and Wang}]{global_hgclr}
Wang, Z.; Wang, P.; Huang, L.; Sun, X.; and Wang, H. 2022.
\newblock Incorporating Hierarchy into Text Encoder: a Contrastive Learning
  Approach for Hierarchical Text Classification.
\newblock In \emph{Proceedings of the 60th Annual Meeting of the Association
  for Computational Linguistics (Volume 1: Long Papers)}, 7109--7119.
  Association for Computational Linguistics.

\bibitem[{Wehrmann, Cerri, and Barros(2018)}]{local_hmcn}
Wehrmann, J.; Cerri, R.; and Barros, R. 2018.
\newblock Hierarchical Multi-Label Classification Networks.
\newblock In Dy, J.; and Krause, A., eds., \emph{Proceedings of the 35th
  International Conference on Machine Learning}, volume~80, 5075--5084. PMLR.

\bibitem[{Wu, Xiong, and Wang(2019)}]{global_l2l}
Wu, J.; Xiong, W.; and Wang, W.~Y. 2019.
\newblock Learning to Learn and Predict: A Meta-Learning Approach for
  Multi-Label Classification.
\newblock In \emph{Proceedings of the 2019 Conference on Empirical Methods in
  Natural Language Processing and the 9th International Joint Conference on
  Natural Language Processing (EMNLP-IJCNLP)}, 4354--4364. Association for
  Computational Linguistics.

\bibitem[{Xu et~al.(2015)Xu, Ba, Kiros, Cho, Courville, Salakhudinov, Zemel,
  and Bengio}]{bigruatt}
Xu, K.; Ba, J.; Kiros, R.; Cho, K.; Courville, A.; Salakhudinov, R.; Zemel, R.;
  and Bengio, Y. 2015.
\newblock Show, Attend and Tell: Neural Image Caption Generation with Visual
  Attention.
\newblock In Bach, F.; and Blei, D., eds., \emph{Proceedings of the 32nd
  International Conference on Machine Learning}, volume~37, 2048--2057. PMLR.

\bibitem[{Yang et~al.(2018)Yang, Sun, Li, Ma, Wu, and Wang}]{global_sgm}
Yang, P.; Sun, X.; Li, W.; Ma, S.; Wu, W.; and Wang, H. 2018.
\newblock SGM: Sequence Generation Model for Multi-label Classification.
\newblock In \emph{Proceedings of the 27th International Conference on
  Computational Linguistics}, 3915--3926. Association for Computational
  Linguistics.

\bibitem[{Yang et~al.(2016)Yang, Yang, Dyer, He, Smola, and Hovy}]{han}
Yang, Z.; Yang, D.; Dyer, C.; He, X.; Smola, A.; and Hovy, E. 2016.
\newblock Hierarchical Attention Networks for Document Classification.
\newblock In \emph{Proceedings of the 2016 Conference of the North American
  Chapter of the Association for Computational Linguistics: Human Language
  Technologies}, 1480--1489. Association for Computational Linguistics.

\bibitem[{Ying et~al.(2021)Ying, Cai, Luo, Zheng, Ke, He, Shen, and
  Liu}]{graphormer}
Ying, C.; Cai, T.; Luo, S.; Zheng, S.; Ke, G.; He, D.; Shen, Y.; and Liu, T.-Y.
  2021.
\newblock Do Transformers Really Perform Badly for Graph Representation?
\newblock In Ranzato, M.; Beygelzimer, A.; Dauphin, Y.; Liang, P.~S.; and
  Vaughan, J.~W., eds., \emph{Advances in Neural Information Processing
  Systems}, volume~34, 28877--28888. Curran Associates, Inc.

\bibitem[{Zhao et~al.(2018)Zhao, Ye, Yang, Lei, Zhang, and
  Zhao}]{global_capsule}
Zhao, W.; Ye, J.; Yang, M.; Lei, Z.; Zhang, S.; and Zhao, Z. 2018.
\newblock Investigating Capsule Networks with Dynamic Routing for Text
  Classification.
\newblock In \emph{Proceedings of the 2018 Conference on Empirical Methods in
  Natural Language Processing}, 3110--3119. Association for Computational
  Linguistics.

\bibitem[{Zhou et~al.(2020)Zhou, Ma, Long, Xu, Ding, Zhang, Xie, and
  Liu}]{global_hiagm}
Zhou, J.; Ma, C.; Long, D.; Xu, G.; Ding, N.; Zhang, H.; Xie, P.; and Liu, G.
  2020.
\newblock Hierarchy-Aware Global Model for Hierarchical Text Classification.
\newblock In \emph{Proceedings of the 58th Annual Meeting of the Association
  for Computational Linguistics}, 1106--1117. Association for Computational
  Linguistics.

\end{thebibliography}

\end{document}